\documentclass[lettersize, 10 pt, conference]{ieeeconf}
\IEEEoverridecommandlockouts                              % This command is only needed if 
\usepackage{amsmath,amsfonts,amssymb}
\usepackage{algorithm}
\usepackage{algorithmic}
\usepackage{array}
\usepackage[caption=false,font=normalsize,labelfont=sf,textfont=sf]{subfig}
\usepackage{textcomp}
\usepackage{stfloats}
\usepackage{url}
\usepackage{verbatim}
\usepackage{graphicx}

\usepackage[pdfborder={0 0 0}]{hyperref}
\usepackage{cleveref}
\crefname{equation}{Eq.}{Eqs.}
\crefname{figure}{Fig.}{Figs.}
\crefname{algorithm}{Alg.}{Algs.}
\usepackage{cite}
\usepackage{bm}
\newcommand{\mbf}[1]{\mathbf{#1}}

\newcommand{\mcal}[1]{\mathcal{#1}}

\usepackage{xcolor}

\usepackage{multirow}
\usepackage{multicol} % \multicolumn{cols}{pos}{text}

\usepackage[table]{xcolor}
\definecolor{bestcolor}{RGB}{255,179,179}     % light red
\definecolor{secondcolor}{RGB}{255,204,153}   % light orange
\definecolor{thirdcolor}{RGB}{255,255,153}    % light yellow
\newcommand{\best}[1]{\cellcolor{bestcolor}\textbf{#1}}
\newcommand{\second}[1]{\cellcolor{secondcolor}#1}
\newcommand{\third}[1]{\cellcolor{thirdcolor}#1}
\usepackage{graphicx} % \resizebox
\usepackage[normalem]{ulem} % double underline
\usepackage{booktabs} % \toprule \midrule

\usepackage{diagbox} % table left-top diagonal line

\usepackage{threeparttable} % table footnote
\definecolor{subcolor}{RGB}{255,179,179}   % red
\definecolor{addcolor}{RGB}{179,255,153}   % green
\newcommand{\add}[1]{\cellcolor{addcolor}#1} % for table ablation study, highlight the increase value
\newcommand{\sub}[1]{\cellcolor{subcolor}#1} % for table ablation study, highlight the decrease value

\usepackage{caption}

\begin{document}

\title{\LARGE \bf
GNFormer: Gaussian Neural Field Transformer for\\
Efficient Camera-based Occupancy Prediction
}

\author{
Guangcheng Chen$^{1,2}$, Li He$^{1,2}$, and Hong Zhang$^{1,2}$, \textit{Life Fellow, IEEE}
\thanks{$^{1}$
	Department of Electronic and Electrical Engineering, Southern University of Science and Technology, China}
\thanks{$^{2}$
	Shenzhen Key Laboratory of Robotics and Computer Vision, Southern University of Science and Technology, Shenzhen, China}
\thanks{$^{*}$
	Corresponding author: Hong Zhang (hzhang@sustech.edu.cn)}   
}

\maketitle
\thispagestyle{empty}
\pagestyle{empty}

\begin{abstract}
Camera-based occupancy prediction provides geometric and semantic scene understanding for robot perception and has been adopted in autonomous vehicle and UAV perception systems.
Gaussian primitives offer a promising alternative by potentially representing only occupied regions, thereby concentrating computation and representational capacity where needed and reducing wasted computation and memory in empty space.
However, existing Gaussian-based methods suffer from poor primitive utilization, as ineffective training objectives leave some Gaussians in empty space, while the limited expressiveness of Gaussian functions restricts their ability to model complex shapes.
To address these limitations, we propose a novel 3D representation, Gaussian Neural Field (GNF), which can accurately map voxel positions to semantic occupancies.
In GNF, Gaussians and a neural network jointly parameterize the field, where the Gaussians define its spatial support and provide features for neural decoding, while the neural network enables more expressive occupancy modeling than Gaussian functions alone.
We further develop the Gaussian Neural Field Transformer (GNFormer), a streamlined and flexible method that predicts GNFs for camera-based occupancy prediction.
Across Occ-ScanNet, SSCBench-KITTI-360, and nuScenes-SurroundOcc benchmarks, GNFormer achieves the highest IoU and mIoU and the fastest inference among the compared single-frame methods.
In the experiments, GNFormer can represent indoor and outdoor scenes with only 256 and 1024 Gaussians, respectively, substantially fewer primitives than prior Gaussian-based methods.
The code will be released.
\end{abstract}


\section{Introduction}

Camera-based occupancy prediction estimates scene geometry and semantics from images~\cite{monoscene2022cvpr,tpvformer2023cvpr}, providing spatial information for robot navigation~\cite{embodiedocc2025iccv,ss2026aaai}.
However, the number of voxels in a dense 3D grid grows rapidly with the spatial range and resolution, imposing substantial computation and memory costs on robotic perception.
Existing methods use sparse voxels~\cite{voxformer2023cvpr}, bird's-eye-view (BEV) features~\cite{flashocc2023}, and tri-perspective-view (TPV) features~\cite{tpvformer2023cvpr} to improve efficiency, yet achieving an accurate and compact semantic occupancy representation remains challenging.

\begin{figure}[t]
	\centering
	\includegraphics[width=\columnwidth]{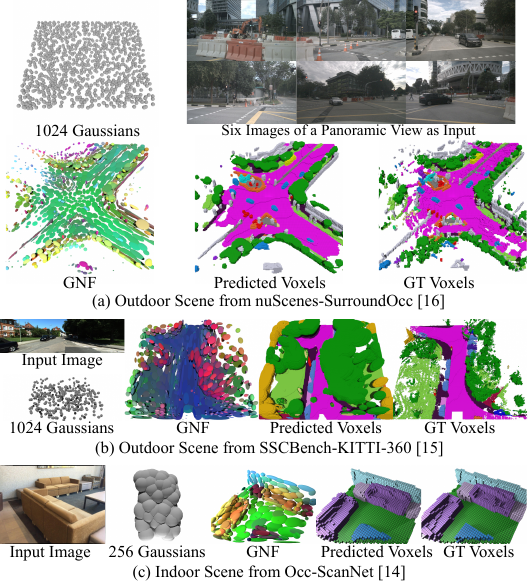}
	\caption{
		GNFormer predicts semantic occupancy from single-frame images and camera parameters using GNF.
		The same architecture can be trained and deployed for different scenarios and camera configurations separately.
	}
	\label{fig:intro}
\end{figure}

Gaussian-based representations provide a promising alternative because each primitive can cover a local spatial neighborhood~\cite{gaussianformer2024eccv,gaussianformer22025cvpr,superocc2026icra,gcchen2026flmocc}.
Existing methods use Gaussian supports to approximate occupancy geometry and their posteriors to aggregate Gaussian semantics~\cite{gaussianformer22025cvpr,gcchen2026flmocc}.
However, existing probabilistic modeling~\cite{gaussianformer22025cvpr,quadformer2025nips,ss2026aaai,s2go2026iclr} may leave some Gaussians under-utilized due to their ineffective training objectives, wasting primitive budget and computation.
Density-based modeling~\cite{gcchen2026flmocc} substantially improves Gaussian utilization.
Yet, the limited expressiveness of simple Gaussian functions restricts the accuracy of the decoded semantic occupancy grids.
Consequently, existing formulations do not simultaneously provide effective Gaussian utilization and expressive voxel-grid modeling.

To address these limitations, we propose Gaussian Neural Field (GNF), a compact semantic occupancy representation that decouples Gaussian spatial allocation from semantic occupancy decoding.
Such decoupling allows us to replace the simple Gaussian function with more expressive functions semantic occupancy decoding.
In GNF, the Gaussians no longer directly determine voxel semantics.
Instead, they specify the spatial range of the neural field, within which semantic occupancy is decoded by an multi-layer perceptron (MLP) given aggregated Gaussian features.
This decoupling allows Gaussians spatial allocation to be optimized for compact coverage of occupied regions, while neural decoding provides the more expressiveness needed for accuracy.

However, cross-entropy supervision for semantics does not constrain the positions of the Gaussians, some Gaussians can remain in empty regions after optimization, wasting primitive capacity and incur unnecessary computation of neural decoding.
We therefore adopt Gaussian regularization~\cite{gcchen2026flmocc}, which is essential for driving Gaussians toward occupied voxels.
By concentrating the spatial support of GNF around occupied regions, it reduces compuation and memory costs while still enjoying the expressive neural decoding for better accuracy.

We further develop Gaussian Neural Field Transformer (GNFormer), a streamlined and flexible feed-forward method that predicts GNF for camera-based occupancy prediction.
GNFormer leverages Transformer~\cite{vaswani2017attention} to transform initial Gaussians with the guidance from single-frame images (captured at the same timestamp synchronously) and camera parameters.
It can be eaisly configurated and trained for indoor and outdoor scenarios and different number of cameras separately.
\Cref{fig:intro} illustrates how the resulting Gaussians parameterize a GNF for semantic occupancy prediction.

We evaluate GNFormer on Occ-ScanNet~\cite{iso2024eccv}, SSCBench-KITTI-360~\cite{li2024sscbench}, and nuScenes-SurroundOcc~\cite{surroundocc2023iccv}.
Across these benchmarks, GNFormer achieves the highest IoU and mIoU and the fastest inference among the compared single-frame methods while using a compact set of Gaussian primitives.
In summary, our contributions are as follows:
\begin{itemize}
\item We propose GNF, a semantic occupancy representation that decouples Gaussian spatial allocation from semantic occupancy decoding, improving accuracy with fewer primitives and lower computation and memory costs.
\item We develop GNFormer, a streamlined and flexible method that predicts GNF for camera-based occupancy prediction.
The same architecture supports different scenarios with different camera configurations.
\item We evaluate GNFormer on one indoor and two outdoor benchmarks with different camera configurations, demonstrating its architectural flexibility, occupancy prediction accuracy, and inference efficiency.
\end{itemize}
\begin{figure*}[t]
	\centering
	\includegraphics[width=\textwidth]{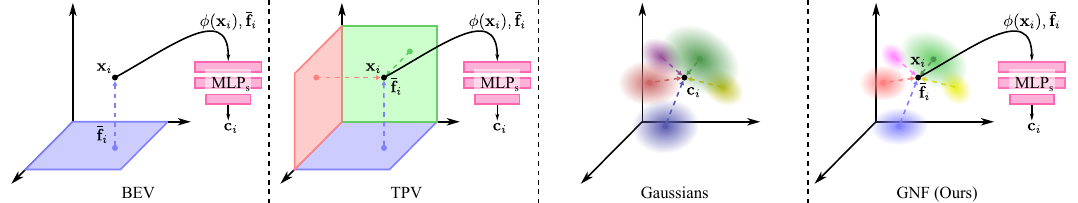}
	\caption{
Decoding semantics from BEV~\cite{bevformer2022} and TPV~\cite{tpvformer2023cvpr} requires traversing the entire space.
Gaussian-based methods~\cite{gaussianformer22025cvpr,gcchen2026flmocc} compute semantics only for regions covered by the Gaussians, using posterior-weighted Gaussian semantics.
GNF further employs a neural decoder for semantic prediction while retaining the efficiency of Gaussian-based representations.
	}
	\label{ch3-rep}
\end{figure*}

\section{Related Work}

\subsection{Gaussian-based Occupancy Prediction}

Gaussian-based methods represent semantic occupancy with a set of geometric primitives~\cite{gaussianformer2024eccv,gaussianformer22025cvpr,ss2026aaai,superocc2026icra,s2go2026iclr,gcchen2026flmocc}.
GaussianFormer~\cite{gaussianformer2024eccv} proposes the first Gaussian-based occupancy prediction framework, but it can still allocate Gaussians to empty regions.
Subsequent methods explore two representative directions: probabilistic modeling, introduced by GaussianFormer-2~\cite{gaussianformer22025cvpr}, and density-based modeling, adopted by FLM-Occ~\cite{gcchen2026flmocc}.
GaussianFormer-2 models voxel occupancy through probabilistic Gaussian composition, and several follow-up works~\cite{quadformer2025nips,ss2026aaai,s2go2026iclr} further develop this formulation.
Across the probabilistic and density-based formulations, Gaussian supports approximate occupancy geometry, while posterior weights derived from their responses aggregate Gaussian semantics into voxel semantics.

However, probabilistic modeling may still leave isolated Gaussians in empty regions, resulting in ineffective primitive utilization.
FLM-Occ attributes this issue to vanishing gradients with respect to Gaussian primitives and instead optimizes their spatial allocation through Gaussian-induced density.
Its likelihood-based regularization encourages Gaussians to concentrate around occupied regions, improving Gaussian allocation.
However, the voxel grid is still decoded directly from Gaussian density and posterior semantics, limiting representational capacity and occupancy accuracy.
Because the spatial supports and semantic posterior weights depend on the same Gaussian parameters, Gaussian allocation remains coupled with semantic decoding.
In contrast, GNF decouples these roles by retaining Gaussian supports for occupancy geometry while replacing posterior semantic aggregation with neural-field decoding.

\subsection{Neural Field-based Occupancy Prediction}
Neural fields provide a continuous mapping from coordinates to scene properties and have gained popularity in 3D vision tasks such as novel view synthesis~\cite{mildenhall2020nerf} and surface reconstruction~\cite{neus2021wang}.
Their neural decoding provides strong capacity for modeling complex geometry and semantic boundaries.
TPVFormer~\cite{tpvformer2023cvpr} combines tri-perspective features with a neural field for fine-grained occupancy prediction.
HybridOcc~\cite{hybridocc2024ral} models binary occupancy with a neural field and supervises it through depth rendering, while subsequent works~\cite{s4c20243dv,scenedino2025iccv,occnerf2025zhang} extend neural fields to semantic occupancy under 2D image supervision.
Despite their expressive decoding, existing neural-field-based methods commonly require dense or grid-structured queries over large spatial regions, limiting their efficiency for 3D occupancy prediction.
GNF addresses this limitation by using Gaussian local supports to define the effective domain of a neural field.
GNF restricts neural decoding to voxels covered by Gaussian supports, where the decoder models semantic occupancy.

\section{Methodology}

\subsection{Overview}
We consider the problem of training an occupancy predictor that takes single-frame input images $\{\mcal{I}_c\}_{c=1}^{N_{\text{cam}}}$ (captured at the same timestamp with one or multiple cameras) with camera parameters and outputs a grid $\{(\mbf{x}_i,\mbf{o}_i)\}_{i=1}^{UVW}$, where $U,V,W$ are the dimensions of the grid, $\mbf{x}_i\in\mathbb{R}^3$ is a voxel center and $\mbf{o}_i\in\mathbb{R}^{N_c}$ is a one-hot vector.
The $N_c$ semantic classes include \emph{empty} and $N_c-1$ object classes.

We first introduce Gaussian Neural Field (GNF), which can fit a semantic occupancy grid with much fewer primitives than voxels (Sec.~\ref{sec:gnf}).
We implement an algorithm to accelerate the conversion (Sec.~\ref{sec:pgf}).
We then design GNFormer that uses a streamlined architecture to predict GNFs given images and camera parameters (Sec.~\ref{sec:arch}).

\subsection{Gaussian Neural Field with Gaussian Regularization}
\label{sec:gnf}
We define a GNF as a function $f_{gn}:\mbf{x}_i\to\mbf{c}_i$ that maps $\mbf{x}_i\in\mathbb{R}^3$ to semantic logits $\mbf{c}_i\in\mathbb{R}^{N_c}$.
$f_{gn}$ consists of a set of 3D Gaussians $\mcal{G}=\{\mbf{g}_j\}_{j=1}^M$ and a multi-layer perceptron, the neural decoder $\operatorname{MLP_s}$.
The number of Gaussians $M$ is a hyperparameter that is adjusted based on the complexity of the scene and the computational resources available.
Each Gaussian $\mbf{g}_j$ has its mean $\bm{\mu}_j\in\mathbb{R}^3$, covariance $\bm{\Sigma}_j\in\mathbb{R}^{3\times 3}$, and feature vector $\mbf{f}_j\in\mathbb{R}^d$.
Given a point $\mbf{x}_i$, its semantic logits $\mbf{c}_i$ is computed as
\begin{equation}\label{eq:agg}
	\bar{\mbf{f}}_i = \sum_{j=1}^M \mcal{K}_{ij} \mbf{f}_j,
\end{equation}
\begin{equation}
	\mbf{c}_i = \operatorname{MLP_s}(\phi(\mbf{x}_i),\bar{\mbf{f}}_i),
\end{equation}
where $\mcal{K}_{ij}=\exp\big(-\frac{1}{2}(\mbf{x}_i-\bm{\mu}_j)^T \bm{\Sigma}_j^{-1} (\mbf{x}_i-\bm{\mu}_j)\big)$ and $\phi$ is the sinusoidal encoding function~\cite{mildenhall2020nerf}.
Therefore, GNF is essentially the function $f_{gn}(\cdot)=\operatorname{MLP_s}(\phi(\cdot),\sum_{j=1}^M 
\mcal{K}_{ij} \mbf{f}_j)$.
To convert a GNF into a semantic occupancy grid, we traverse every voxel and query the GNF with the center of the voxel to get the semantics.

Different from existing Gaussian-based methods that use $\mbf{c}_i=(\sum_{j=1}^M \mcal{K}_{ij}\mbf{f}_j)/(\sum_{j=1}^M \mcal{K}_{ij})$ and $\mbf{f}_j\in\mathbb{R}^{N_c}$ to calculate the semantics,
GNF decouples the spatial allocation of the Gaussians from semantic occupancy decoding and leverages the expressive representation capacity of $\operatorname{MLP_s}$.

Moreover, since the norm of $\mcal{K}_{ij}\mbf{f}_j$ decreases exponentially as the distance between $\mbf{x}_i$ and $\bm{\mu}_j$ increases, we further simplify \cref{eq:agg} by omitting the small terms and get:
\begin{equation}\label{eq:agg_local}
	\bar{\mbf{f}}_i = \sum_{j\in\mcal{N}_i} \mcal{K}_{ij} \mbf{f}_j,
\end{equation}
where $\mcal{N}_i$ is the index set of the Gaussians whose responses at the $i$-th voxel center exceed a threshold.
Specifically, we define each Gaussian's local support by thresholding its Mahalanobis distance at $\tau$, i.e., $\mcal{K}_{ij}>\tau$ and include only Gaussians whose local supports contain the center of a voxel when computing the semantics of the voxel.
Since the Gaussians now explicitly define the spatial support of a GNF,
one can skip regions that are out of the spatial support when converting a GNF into a semantic occupancy grid, which is critical for computational efficiency (see \cref{tab:aggregation-efficiency}).

\begin{figure}[t]
	\centering
	\includegraphics[width=\columnwidth]{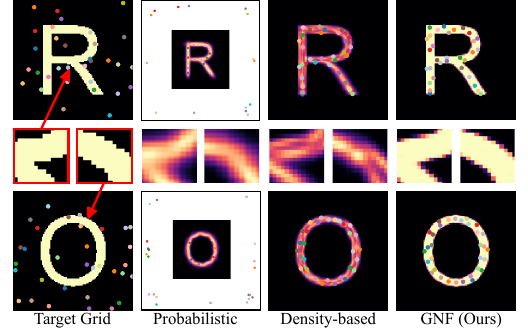}
	\caption{
		Fitting 2D ``R'' and ``O'' occupancy grids (100$\times$100) with 32 Gaussians using different methods.
		Gaussian centers are shown over the grids with their initial positions shown in the first column.
		Only Gaussian centers are shown for clarity, and brighter pixels indicate occupied regions.
		GNF not only drives all Gaussians towards occupied regions but also fits sharp boundaries with an extra MLP.
	}
	\label{fig:2D_exp}
\end{figure}

The parameters of a GNF, i.e., those of $\mcal{G}$ and $\operatorname{MLP_s}$, can be optimized by minimizing the cross-entropy loss (CE):
\begin{equation}\label{loss:ce}
	\mcal{L}_{\text{CE}} = \frac{1}{UVW}\sum_{i=1}^{UVW}\operatorname{CE}(\hat{\mbf{o}}_{i}, \mbf{o}_{i}^{gt}),
	\ \hat{\mbf{o}}_{i} = \operatorname{Softmax}(\mbf{c}_i),
\end{equation}
where $\mbf{o}_{i}^{gt}$ is a ground-truth one-hot vector.
Ideally, the Gaussians of a GNF are expected to concentrate around occupied regions, thereby reducing the computational cost and representation capacity on empty regions.
However, some Gaussians can stay in empty regions after optimization, because of the absence of direct constraints on the allocation of Gaussians.
Therefore, we add a Gaussian regularization~\cite{gcchen2026flmocc} (GR) to drive the Gaussians towards occupied voxels:
\begin{equation}\label{eq:flm_loss}
    \mcal{L}_{\text{GR}} = \log\sum_{j=1}^M v_j - \frac{1}{|\mcal{X}^{\text{obj}}|}\sum_{i=1}^{|\mcal{X}^{\text{obj}}|} \log\sum_{j\in\mathcal{N}_i} \mcal{K}_{ij},
\end{equation}
where $\mcal{X}^{\text{obj}}\subset\mcal{X}^{\text{full}}=\{\mbf{x}_i\}^{UVW}_{i=1}$ is the object voxel set of the ground-truth occupancy grid, and $v_j=\int_{\mathbb{R}^3} \mcal{K}(\mbf{x}|\bm{\mu}_j, \bm{\Sigma}_j) d\mbf{x}$ is the volume of the $j$-th Gaussian.
The final optimization objective is a weighted sum of the two losses:
\begin{equation}\label{eq:loss_total}
    \mcal{L}_{\text{total}} = \mcal{L}_{\text{CE}} + \lambda \mcal{L}_{\text{GR}}.
\end{equation}

As shown in \Cref{fig:2D_exp}, we conduct a 2D pilot experiment to study existing Gaussian-based representations.
We optimize 32 Gaussians to fit target occupancy grids with the optimization objectives from different occupancy modelings.
We compare GNF with the probabilistic modeling introduced by GaussianFormer-2~\cite{gaussianformer22025cvpr} and the density-based modeling of FLM-Occ~\cite{gcchen2026flmocc}.
The probabilistic modeling leaves many Gaussians under-utilized because the occupied-voxel gradients that attract Gaussians can vanish~\cite{gcchen2026flmocc}.
The empty-voxel gradients then push these Gaussians away from the grid until their responses become negligible.
The FLM loss~\cite{gcchen2026flmocc} improves Gaussian allocation by encouraging the primitives to cover occupied voxels, but its unbounded density requires a non-trivial threshold and cannot reliably recover the target occupancy grid.
In contrast, GNF separates Gaussian spatial allocation from semantic occupancy decoding, combining effective Gaussian utilization with accurate occupancy fitting through neural-field decoding.
This combination enables accurate occupancy prediction with fewer Gaussians, reducing computation and memory demands for robotic perception.

\subsection{Projection-based Point--Gaussian Pair Filtering}
\label{sec:pgf}
To efficiently aggregate Gaussian features in \cref{eq:agg_local} and densities for Gaussian regularization in \cref{eq:flm_loss}, we propose a projection-based point--Gaussian filtering strategy to solve $\mathcal{N}_i$.
This procedure accelerates the enumeration of local point--Gaussian pairs without changing the GNF representation.
Given a set of 3D directions $\left\{ \mbf{d}_l\in \mathbb{S}^{2}\right\}_{l=1}^{L}$,
we project each point $\mbf{x}_i$ and each Gaussian $\mbf{g}_j$ onto
1D subspaces:
\begin{equation}
	\bar{x}_{i,l}
	=
	\mbf{d}_l^{\top} \mbf{x}_i,
\end{equation}
\begin{equation}
	\bar{\mu}_{j,l}
	=
	\mbf{d}_l^{\top} \bm{\mu}_j,
	\quad
	\bar{\sigma}_{j,l}^{2}
	=
	\mbf{d}_l^{\top}
	\bm{\Sigma}_j
	\mbf{d}_l .
\end{equation}
In our experiments, we use the three canonical axes as the projection directions, which suffice to capture the spatial support of a 3D Gaussian.
We then define an effective interval for each Gaussian and each projection direction.
Here, $\tau>0$ is a support-radius hyperparameter that specifies the width of the local support in units of the projected standard deviation;
a larger $\tau$ keeps more point--Gaussian candidates, while a smaller $\tau$ produces a tighter local support:
\begin{equation}\label{eq:interval}
	S_{j,l}
	=
	\left[
	\bar{\mu}_{j,l}
	-
	\tau
	\bar{\sigma}_{j,l},
	\;
	\bar{\mu}_{j,l}
	+
	\tau
	\bar{\sigma}_{j,l}
	\right].
\end{equation}
A pair $(i,j)$ that pairs the $i$-th point and the $j$-th Gaussian is retained as a candidate only if the projected
point lies inside the corresponding interval for all selected directions:
\begin{equation}
	\label{eq:candidate}
	(i,j) \in \mcal{C}
	\ \Longleftrightarrow \ 
	\left|
	\bar{x}_{i,l}
	-
	\bar{\mu}_{j,l}
	\right|
	\leq
	\tau
	\bar{\sigma}_{j,l},
	\ 
	\forall l = 1,\dots,L .
\end{equation}

Instead of naively testing all $P\times M$ pairs with \cref{eq:candidate}, we design an algorithm in \Cref{alg:pgf} to find the point index ranges that are covered by the projected 1D Gaussians.
This reduces the time and space complexities from $\mathcal{O}(PM)$ to $\mathcal{O}(M\log M + M\log P + K)$ and $\mathcal{O}(P + M + K)$, respectively.
All the operations in \Cref{alg:pgf} are natively supported by $\texttt{PyTorch}$ and can leverages GPU for acceleration.
We also implement $\texttt{CUDA}$ kernels to obtain more efficiency and further reduce memory consumption.
This is critical for efficiency when large $P$ and $M$ are large.

To further reduce point-Gaussian pairs, we additionally apply a Mahalanobis-distance test to retain pairs that satisfy
\begin{equation}
	(\mbf{x}_i - \bm{\mu}_j)^{\top}
	\bm{\Sigma}_j^{-1}
	(\mbf{x}_i - \bm{\mu}_j)
	\leq
	\tau^2 .
\end{equation}

\begin{algorithm}[t]
	\caption{Parallel 1D Point--Gaussian Pairing}
	\label{alg:pgf}
	\begin{algorithmic}[1]
		\item[] \hspace{-\algorithmicindent}\textbf{Input:}
		projected points $\bar{\mbf{x}}\in\mathbb{R}^{P}$,
		projected means $\bar{\bm{\mu}}\in\mathbb{R}^{M}$,
		support radii $\tau\bar{\bm{\sigma}}\in\mathbb{R}^{M}$
		\item[] \hspace{-\algorithmicindent}\textbf{Output:}
		point indices $\mbf{i}\in\mathbb{R}^K$,
		Gaussian indices $\mbf{j}\in\mathbb{R}^K$
		\STATE $\mbf{a}\leftarrow\operatorname{argsort}(\bar{\mbf{x}})$,
		$\bar{\mbf{x}}^{s}\leftarrow\bar{\mbf{x}}[\mbf{a}]$
		\STATE
		$\mbf{l}\leftarrow\bar{\bm{\mu}}-\tau\bar{\bm{\sigma}}$,
		$\mbf{r}\leftarrow\bar{\bm{\mu}}+\tau\bar{\bm{\sigma}}$
		\STATE $\mbf{s}\leftarrow\operatorname{searchsorted}
		(\bar{\mbf{x}}^{s},\mbf{l},\mathrm{left})$
		\STATE $\mbf{e}\leftarrow\operatorname{searchsorted}
		(\bar{\mbf{x}}^{s},\mbf{r},\mathrm{right})$
		\STATE $\mbf{c}\leftarrow\mbf{e}-\mbf{s}$,
		\ $K\leftarrow\sum_{j=1}^{M}c_j$
		\STATE $\mbf{j}\leftarrow\operatorname{repeat\_interleave}([1,\ldots,M],\mbf{c})$
		\STATE $\bm{\rho}\leftarrow\operatorname{exclusive\_prefix\_sum}(\mbf{c})$
		\STATE $\mbf{o}\leftarrow[0,\ldots,K-1]
		-\operatorname{repeat\_interleave}(\bm{\rho},\mbf{c})$
		\STATE $\mbf{h}\leftarrow
		\operatorname{repeat\_interleave}(\mbf{s},\mbf{c})+\mbf{o}$
		\STATE $\mbf{i}\leftarrow\mbf{a}[\mbf{h}]$
		\RETURN $\mbf{i},\mbf{j}$
	\end{algorithmic}
\end{algorithm}

\subsection{Architecture of Gaussian Neural Field Transformer}
\label{sec:arch}
\begin{figure*}[t]
	\centering
 	\includegraphics[width=\textwidth]{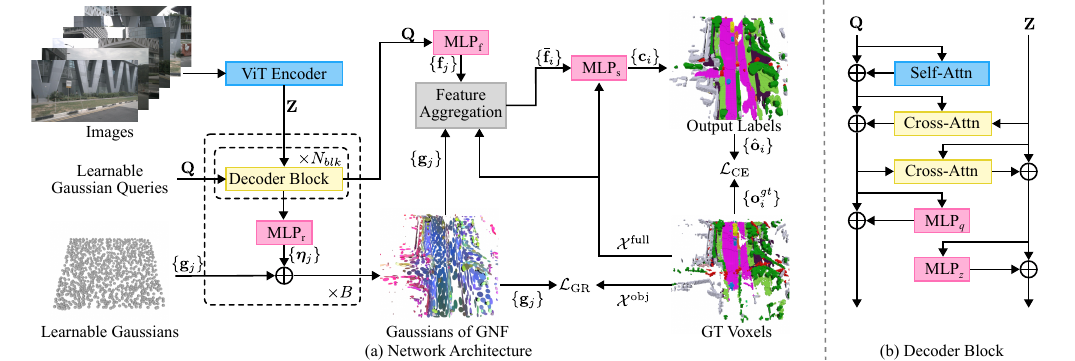}
	\caption{
(a) Given input images and camera parameters, GNFormer employs Transformer blocks~\cite{vaswani2017attention} to transform a set of Gaussians and queries into a GNF.
During training, output semantics are supervised using cross-entropy loss and Gaussians are regularized following FLM-Occ~\cite{gcchen2026flmocc}.
Note that the camera parameters and positional encoding are omitted for brevity.
(b) Each decoder block adds an extra cross-attention layer to enable information exchange among all tokens of $\mbf{Q}$ and $\mbf{Z}$.
	}
	\label{fig:arch}
\end{figure*}

Given the proposed GNF, we design GNFormer, an end-to-end occupancy prediction method that can be trained for predicting GNFs given images and camera parameters.
GNFormer has a Transformer architecture~\cite{vaswani2017attention} as shown in \Cref{fig:arch} (a).
We design the architecture so that it can be easily configured for a specific scenario and number of cameras before training.
Given input images $\{\mcal{I}_c\}_{c=1}^{N_{\text{cam}}}$ at one timestamp, we first use a ViT~\cite{dinov22023} to encode the images:
\begin{equation}
    \mbf{Z}
    =
    \operatorname{Concat}(\{\operatorname{ViT}(\mcal{I}_c)\}_{c=1}^{N_{\text{cam}}}),
\end{equation}
where $\mbf{Z}\in\mathbb{R}^{(N_\text{cam}\times N_\text{tok}) \times d_h}$ denotes the concatenated image
tokens from all cameras and $N_{\text{tok}}$ is the number of tokens per image.
The decoder refines a set of learnable Gaussians:
\begin{equation}
    \mcal{G}^{0}
    =
    \{(\mbf{g}^0_j,\mbf{q}^{0}_{j})\}_{j=1}^{M},
    \ \mbf{g}^0_j=[\bm{\mu}^{0}_{j}; \mbf{s}_j^0; \mbf{r}^{0}_{j}],
\end{equation}
where $\bm{\mu}^{0}_{j}\in\mathbb{R}^{3}$ is the initial mean, $\mbf{s}_j^0\in\mathbb{R}^{3}$ and $\mbf{r}^{0}_{j}\in\mathbb{R}^{4}$ are the scale and rotation parameters, respectively, and define the covariance $\bm{\Sigma}=f_{R}(\mbf{r})\mbf{s}\mbf{s}^{\top}f_{R}^{\top}(\mbf{r})$, $f_{R}(\cdot)$ converts a quaternion to a rotation matrix, and $\mbf{q}^{0}_{j}\in\mathbb{R}^{d_h}$ is the corresponding Gaussian query.
These learnable Gaussians are initialized randomly within the grid boundaries before training.
Once the training is done, their values are frozen and shared for any input images.

The image tokens $\mbf{Z}$ serve as keys and values for the decoder.
The decoder contains $N_{dec}$ Transformer blocks and progressively refines the Gaussians.
For every $N_{blk}$ blocks, each updated query $\mbf{q}_{j}^{t-1}$ is
decoded into a residual update:
\begin{equation}\label{eq:update}
    \bm{\eta}_{j}^{t}
    =
    \operatorname{MLP}_\text{r}(\mbf{q}_{j}^{t-1}),
    \ \mbf{g}^{t}_j=\mbf{g}^{t-1}_j+\bm{\eta}_{j}^{t}
\end{equation}
and $t$ denotes the number of refinements.
Note that in our actual implementation, $\bm{\mu}^t_{j}$ and $\mbf{s}_j^t$ are updated in $\operatorname{asinh}$ space and $\log$ space, respectively, and $\mbf{r}^t_{j}$ is normalized after the final update.
We use the decoder to refine rather than directly regress the Gaussians because
the refinement process simplifies the Gaussian prediction task and stabilizes training~\cite{gaussianformer2024eccv}, leading to better performance.
The final queries are mapped to compact Gaussian features:
\begin{equation}
    \mbf{f}_{j}
    =
    \operatorname{MLP}_\text{f}(\mbf{q}^{B}_{j}),
    \ B=N_{dec}/N_{blk}.
\end{equation}
Therefore, the final Gaussian set is
\begin{equation}
    \mcal{G}^B
    =
    \{\mbf{g}_{j}^B\}_{j=1}^{M},
    \quad
    \mbf{g}_{j}^B
    =
    (\bm{\mu}_{j}^B,\mbf{s}_j^B,\mbf{r}_j^B,\mbf{f}_{j}).
\end{equation}

To enable information exchanges among all tokens for better performance, including the queries $\mbf{Q}$ and the image tokens $\mbf{Z}$, while reducing the computational cost of decoding from dense image tokens,
we adopt the bidirectional cross-attention design of LagerNVS~\cite{szymanowicz2026lagernvs} as shown in \Cref{fig:arch} (b).
The design has much lower computational complexity than self-attention over all tokens.

To provide geometric guidance for the attention layers in the decoder blocks, we inject geometry through MRoPE~\cite{mrope2026iclr}.
In self-attention, each Gaussian mean $\bm{\mu}_{j}$ is used as a 3D RoPE coordinate.
In cross-attention, we use a ray-based 6D RoPE representation.
For each Gaussian, we construct
\begin{equation}
    \mbf{r}^{G}_{j}
    =
    \left[
        \frac{\bm{\mu}_{j}}{\|\bm{\mu}_{j}\|_2+\epsilon};
        [0,0,0]^\top
    \right],
\end{equation}
where the first term is the normalized direction from the ego origin to the
Gaussian center, and the second term, $\mbf{t}_{0}=[0,0,0]^\top$, is a zero translation vector.
For an image token, we use the camera parameters to construct a ray representation:
\begin{equation}
    \mbf{r}^{I}_{c,k}
    =
    \left[
        \mbf{d}_{c,k};
        \mbf{t}_{c}
    \right],
\end{equation}
where $\mbf{d}_{c,k}$ is the ray direction of the $k$-th image patch center in the $c$-th camera, and $\mbf{t}_{c}$ is the corresponding camera origin.

Similar to fitting a GNF to an occupancy grid, GNFormer is trained by minimizing $\mathcal{L}_\text{total}$ in \cref{eq:loss_total}.
For efficiency, when computing $\mathcal{L}_\text{CE}$, we use all object voxels and randomly sample empty voxels at four times the number of object voxels.
Note that $\operatorname{MLP_s}$ is a part of GNFormer here and thus is shared for any inputs.
GNFormer predicts Gaussians $\mcal{G}^B$ of GNFs for different input images and camera parameters.

\section{Experiments}

In this section, we train GNFormer on three datasets seperately and evaluate its semantic occupancy prediction performance~\cref{exp:datasets,exp:imple,exp:results}.
To investigate the factors that affect the performance of GNFormer, we provide ablation studies on GNFormer architecture and the ViT~\cite{dinov22023} initialization using the nuScenes-SurroundOcc~\cite{surroundocc2023iccv} dataset.

\begin{table}[t] %
    \centering
    \footnotesize
    \setlength{\tabcolsep}{0.005\linewidth}   
    \renewcommand\arraystretch{1.05}
    \caption{
        % \textbf{
            Summary of Datasets and Hyperparameters
        % }
    }
    \label{tab:datasets}
\begin{tabular}{c|l|c|c|c}
\toprule
&\diagbox{Param}{Dataset} & Occ-ScanNet & SSCBench & nuScenes \\
\midrule
\multirow{5}{*}{\rotatebox{90}{Statistics}}
&Train/Val/Test & 45610/19705/-- & 8487/1812/2566 & 28130/6019/-- \\
&Num. Views & 1 & 1 & 6 \\
&Grid Size & $60^2 \times 32$ & $256^2\times 32$ & $200^2\times 16$ \\
&Voxel Size & 0.08 m & 0.2 m & 0.5 m \\
&\#Obj. Cls. & 11 & 18 & 16 \\
\midrule
\multirow{5}{*}{\rotatebox{90}{Hyperparameters}}
&Input Res. & $518\times686$ & $336\times 1218$ & $420\times742$\\
&ViT Variant& B & L & L\\
&Gau. Init. Scale & 0.5 m & 1.0 m & 1.0 m \\
&Batch Size & 64 & 32 & 32 \\
&Warmup Iters & 1000 & 500 & 1000 \\
&Max Iters & 40k & 10k & 25k \\

\bottomrule 
\end{tabular}
\end{table}

\subsection{Datasets, Compared Methods, and Evaluation Metrics}\label{exp:datasets}
We train and evaluate three GNFormers with three configurations on three datasets:
Occ-ScanNet~\cite{iso2024eccv} for indoor single-view settings, SSCBench-KITTI-360~\cite{li2024sscbench} for outdoor monocular settings, and nuScenes-SurroundOcc~\cite{surroundocc2023iccv} for outdoor surround-view settings.
Their statistics are in \Cref{tab:datasets}.

Most occupancy prediction methods are designed and evaluated for particular scene domains and camera configurations, including indoor single-view and outdoor monocular or surround-view settings.
Consequently, the sets of compared methods differ across tables, with limited overlap between indoor and outdoor methods.
We use intersection-over-union (IoU) for geometry, mean IoU (mIoU) and class-wise IoU for semantics, and frames per second (FPS) for efficiency.
\definecolor{ceiling}{RGB}{214,  38, 40}   %
\definecolor{floor}{RGB}{43, 160, 4}     %
\definecolor{wall}{RGB}{158, 216, 229}  %
\definecolor{window}{RGB}{114, 158, 206}  %
\definecolor{chair}{RGB}{204, 204, 91}   %
\definecolor{bed}{RGB}{255, 186, 119}  %
\definecolor{sofa}{RGB}{147, 102, 188}  %
\definecolor{table}{RGB}{30, 119, 181}   %
\definecolor{tvs}{RGB}{160, 188, 33}   %
\definecolor{furniture}{RGB}{255, 127, 12}  %
\definecolor{objects}{RGB}{196, 175, 214} %

\begin{table*}[t]
	\centering
	\footnotesize
	\setlength{\tabcolsep}{0.006\linewidth}
	\renewcommand\arraystretch{1.0}
	\caption{
		% \textbf{
			Results on Occ-ScanNet~\cite{iso2024eccv} Validation Set
		% }
	}
	\label{tab:exp_occscannet}
%	\resizebox{\textwidth}{!}{
\begin{threeparttable}
		\begin{tabular}{lr|cc|cc|ccccccccccc|c}
			\toprule 
			Method & Venue & Representation & Number
			& IoU & mIoU
			& \rotatebox{90}{\textcolor{ceiling}{$\blacksquare$} Ceiling} 
			& \rotatebox{90}{\textcolor{floor}{$\blacksquare$} Floor}
			& \rotatebox{90}{\textcolor{wall}{$\blacksquare$} Wall} 
			& \rotatebox{90}{\textcolor{window}{$\blacksquare$} Window} 
			& \rotatebox{90}{\textcolor{chair}{$\blacksquare$} Chair} 
			& \rotatebox{90}{\textcolor{bed}{$\blacksquare$} Bed} 
			& \rotatebox{90}{\textcolor{sofa}{$\blacksquare$} Sofa} 
			& \rotatebox{90}{\textcolor{table}{$\blacksquare$} Table} 
			& \rotatebox{90}{\textcolor{tvs}{$\blacksquare$} TVs} 
			& \rotatebox{90}{\textcolor{furniture}{$\blacksquare$} Furniture} 
			& \rotatebox{90}{\textcolor{objects}{$\blacksquare$} Objects} 
%			& \rotatebox{90}{Ceiling} 
%			& \rotatebox{90}{Floor}
%			& \rotatebox{90}{Wall} 
%			& \rotatebox{90}{Window} 
%			& \rotatebox{90}{Chair} 
%			& \rotatebox{90}{Bed} 
%			& \rotatebox{90}{Sofa} 
%			& \rotatebox{90}{Table} 
%			& \rotatebox{90}{TVs} 
%			& \rotatebox{90}{Furniture} 
%			& \rotatebox{90}{Objects} 
			& FPS
			\\
			
			\midrule
			MonoScene~\cite{monoscene2022cvpr}  & CVPR22 & Dense Voxels & 129.6K  & 41.6  & 24.6 & 15.2 & 44.7 & 22.4 & 12.6 & 26.1 & 27.0 & 35.9 & 28.3 & 6.6 & 32.2 & 19.8 & --- \\
			
			ISO~\cite{iso2024eccv}  & ECCV24 & Dense Voxels & 129.6K  & 42.2 & 28.7 & 19.9 & 41.9 & 22.4 & 17.0 & 29.0 & 42.4 & 42.0 & 29.6 & 10.6 & 36.4 & 24.6 & --- \\
			
			DiScene~\cite{discene2025ral}  & RAL25 & Gaussians & 15.4K & 52.0 & 47.2 & 45.3 & 50.6 & 40.4 & 36.7 & 42.3 & 59.7 & 62.0 &45.6 & 41.2 & 52.4 & 42.7 & 11.0 \\
			
			EmbodiedOcc~\cite{embodiedocc2025iccv}  & ICCV25 & Gaussians  & 16.2K & 53.6  & 45.2 & 39.6 & 50.4 & 41.4 & 31.7 & 40.9 & 55.0 & 61.4 & 44.0 & 36.1 & 53.9 & 42.2  & \ 7.8\\ % ICCV data

			SplatSSC~\cite{ss2026aaai}  & AAAI26 & Gaussians & 1200  & \third{62.8} & \third{51.8} & \third{49.1} & \third{59.0} & \third{48.3} & \third{38.8} & \third{47.4} & \third{62.4} & \third{67.0}
			& \third{49.5} & \third{42.6} & \third{60.7} & \third{45.4} & \ \third{8.7} \\

			FLM-Occ~\cite{gcchen2026flmocc}
			& ECCV26 & Superquadrics & 1024
			& \second{71.7} & \second{64.0}  & \second{62.5} & \second{70.6} & \second{58.0} & \second{53.1} & \second{58.8}
			& \second{74.0} & \second{77.9} & \second{63.5} & \second{55.2} & \second{70.8}
			& \second{60.3} & \second{25.4}\\
	
			\cmidrule(lr){1-18}
			
			GNFormer (ours)
 & Ours & GNF & 256
& \best{74.3} & \best{66.9} & \best{63.5} & \best{73.5} & \best{60.1} & \best{54.9}
& \best{62.8} & \best{77.1} & \best{81.0} & \best{68.1} & \best{57.6}
& \best{73.6} & \best{63.4} & \best{30.8}\\

			\bottomrule
		\end{tabular}
%	}
The best, second-best, and third-best results are highlighted in red, orange, and yellow, respectively.
FPS results are measured on a single RTX 3090 GPU.
For GNF, ``Number'' denotes the number of used Gaussians (same notation hereafter).
\end{threeparttable}
\end{table*}

\definecolor{carcolor}{RGB}{111,149,238}
\definecolor{bicyclecolor}{RGB}{136,227,242}
\definecolor{motorcyclistcolor}{RGB}{138, 42, 90}
\definecolor{truckcolor}{RGB}{74,32,172}
\definecolor{othervehiclecolor}{RGB}{96,81,242}
\definecolor{personcolor}{RGB}{234, 60, 49}
\definecolor{roadcolor}{RGB}{234,51,246}
\definecolor{parkingcolor}{RGB}{241,156,249}
\definecolor{sidewalkcolor}{RGB}{68,8,72}
\definecolor{othergroundcolor}{RGB}{160,32,76}
\definecolor{motorcyclecolor}{RGB}{37,59,146}
% \definecolor{trunkcolor}{RGB}{126, 65, 22}
% \definecolor{bicyclistcolor}{RGB}{234, 66, 195}
\definecolor{buildingcolor}{RGB}{246,202,69}
\definecolor{fencecolor}{RGB}{238, 128, 69}
\definecolor{vegetationcolor}{RGB}{79, 173, 50}
\definecolor{terraincolor}{RGB}{171, 238, 105}
\definecolor{polecolor}{RGB}{252, 241, 161}
\definecolor{trafficsigncolor}{RGB}{233, 51, 35}
\definecolor{other-struct.color}{RGB}{255, 150, 0}
\definecolor{other-objectcolor}{RGB}{50, 255, 255}

\begin{table*}[t] %
	\centering
	\footnotesize
	\setlength{\tabcolsep}{0.003\linewidth}   
	\renewcommand\arraystretch{1.0}
	\caption{
		% \textbf{
			Results on SSCBench-KITTI-360~\cite{li2024sscbench} Test Set
		% }
	}
	\label{tab:exp_kitti}
	\begin{threeparttable}
%	\resizebox{1\linewidth}{!}{
		\begin{tabular}{lr|cc|cc|cccccccccccccccccccc}
			\toprule
			Method 
			& Venue & Represent. & Number
			& IoU
			& mIoU
			&\rotatebox{90}{\textcolor{carcolor}{$\blacksquare$} car}
			&\rotatebox{90}{\textcolor{bicyclecolor}{$\blacksquare$} {bicycle}}
			&\rotatebox{90}{\textcolor{motorcyclecolor}{$\blacksquare$} {motorcycle}}
			&\rotatebox{90}{\textcolor{truckcolor}{$\blacksquare$} {truck}}
			&\rotatebox{90}{\textcolor{othervehiclecolor}{$\blacksquare$} {other-veh.}}
			&\rotatebox{90}{\textcolor{personcolor}{$\blacksquare$} {person}}
			&\rotatebox{90}{\textcolor{roadcolor}{$\blacksquare$} {road}}  
			&\rotatebox{90}{\textcolor{parkingcolor}{$\blacksquare$} {parking}}
			&\rotatebox{90}{\textcolor{sidewalkcolor}{$\blacksquare$} {sidewalk}}
			&\rotatebox{90}{\textcolor{othergroundcolor}{$\blacksquare$} {other-grnd}}
			&\rotatebox{90}{\textcolor{buildingcolor}{$\blacksquare$} {building}}
			&\rotatebox{90}{\textcolor{fencecolor}{$\blacksquare$} {fence}}
			&\rotatebox{90}{\textcolor{vegetationcolor}{$\blacksquare$} {vegetation}}
			&\rotatebox{90}{\textcolor{terraincolor}{$\blacksquare$} {terrain}}
			&\rotatebox{90}{\textcolor{polecolor}{$\blacksquare$} {pole}}
			&\rotatebox{90}{\textcolor{trafficsigncolor}{$\blacksquare$} {traf.-sign}}
			&\rotatebox{90}{\textcolor{other-struct.color}{$\blacksquare$} {other-struct.}}
			&\rotatebox{90}{\textcolor{other-objectcolor}{$\blacksquare$} {other-object}}
%			&\rotatebox{90}{car}
%			&\rotatebox{90}{bicycle}
%			&\rotatebox{90}{motorcycle}
%			&\rotatebox{90}{truck}
%			&\rotatebox{90}{other-veh.}
%			&\rotatebox{90}{person}
%			&\rotatebox{90}{road}  
%			&\rotatebox{90}{parking}
%			&\rotatebox{90}{sidewalk}
%			&\rotatebox{90}{other-grnd}
%			&\rotatebox{90}{building}
%			&\rotatebox{90}{fence}
%			&\rotatebox{90}{vegetation}
%			&\rotatebox{90}{terrain}
%			&\rotatebox{90}{pole}
%			&\rotatebox{90}{traf.-sign}
%			&\rotatebox{90}{other-struct.}
%			&\rotatebox{90}{other-object}
			
			\\
			% \\\midrule

			\midrule
			MonoScene~\cite{monoscene2022cvpr} & CVPR22 & Dense Vox & 2.1M & {37.9} & {12.3} & {19.3} & {\ 0.4} & {0.6} & {8.0} & {2.0} & {0.9} & {48.4} & {11.4} & {28.1} & {3.2} & {32.9} & {3.5} & {26.2} & {16.8} & \third{6.9} & {5.7} & {4.2} & {3.1}\\

			TPVFormer~\cite{tpvformer2023cvpr}         & CVPR23 & Planes & 3 & {40.2} & {13.6} & {21.6} & {1.1} & {1.4} & {8.1} & {2.6} & {2.4} & {53.0} & {12.0} & {31.1} & \third{3.8} & {34.8} & {4.8} & {30.1} & {17.5} & \second{7.5} & {5.9} & {5.5} & {2.7}\\

			OccFormer~\cite{occformer2023iccv}         & ICCV23 & Dense Vox & 2.1M & \third{40.3} & {13.8} & \third{22.6} & {0.7} & {0.3} & {9.9} & {3.8} & \third{2.8} & \second{54.3} & \second{13.4} & {31.5} & {3.6} & \second{36.4} & {4.8} & \third{31.0} & \third{19.5} & \best{7.8} & \second{8.5} & \third{7.0} & \third{4.6}\\
			
			GaussianFormer~\cite{gaussianformer2024eccv} & ECCV24 & Gaussians & 38.4K & 35.4 & {12.9} & 18.9 & {1.0} & \best{4.6} & \second{18.1} & \second{7.6} & \second{3.4} & 45.5 & 10.9 & 25.0 & \best{5.3} & 28.4 & \second{5.7} & {29.5} & 8.6 & 3.0 & 2.3 & \best{9.5} & \second{5.1} \\
			
			GaussianFormer-2~\cite{gaussianformer22025cvpr} & CVPR25 & Gaussians & 38.4K & 38.4 & \third{13.9} & 21.1 & \best{2.6} & \second{4.2} & 12.4 & \third{5.7} & 1.6 & \third{54.1} & 11.0 & \third{32.3} & 3.3 & 32.0 & 5.0 & 28.9 & 17.3 & 3.6 & 5.5 & 5.9 & 3.5\\

			S2GO-Base*~\cite{s2go2026iclr} & ICLR26 & Gaussians & 3600 & \second{40.8} & \second{15.1}  & \second{22.7} & \third{1.3} & 1.7 & \third{15.9} & 5.1 & 2.1 & 53.8 & \third{13.3}
			& \second{33.4} & \third{3.8}  & \third{35.3} & \best{7.2} & \second{31.2} & \second{21.1} & 6.4 & \third{6.5} & 6.0 & 4.2 \\
			\midrule
			GNFormer (ours)& Ours & GNF & 1024 & \best{42.5} & \best{17.1} & \best{23.8} & \second{1.5} & \third{3.4} & \best{20.9} & \best{9.6} & \best{3.7} & \best{59.3} & \best{16.5} & \best{35.8} & \second{4.8} & \best{37.0} & \third{5.6} & \best{32.9} & \best{21.8} & {6.7} & \best{10.0}& \second{8.3} & \best{5.3}\\
			\bottomrule
		\end{tabular}
%	}
{
* denotes methods that use sequential images as input and leverage streaming model structures for efficiency; all others use single-frame input.
}
\end{threeparttable}
\end{table*}

\definecolor{nbarrier}{RGB}{255, 120, 50}
\definecolor{nbicycle}{RGB}{255, 192, 203}
\definecolor{nbus}{RGB}{255, 255, 0}
\definecolor{ncar}{RGB}{0, 150, 245}
\definecolor{nconstruct}{RGB}{0, 255, 255}
\definecolor{nmotor}{RGB}{200, 180, 0}
\definecolor{npedestrian}{RGB}{255, 0, 0}
\definecolor{ntraffic}{RGB}{255, 240, 150}
\definecolor{ntrailer}{RGB}{135, 60, 0}
\definecolor{ntruck}{RGB}{160, 32, 240}
\definecolor{ndriveable}{RGB}{255, 0, 255}
\definecolor{nother}{RGB}{139, 137, 137}
\definecolor{nsidewalk}{RGB}{75, 0, 75}
\definecolor{nterrain}{RGB}{150, 240, 80}
\definecolor{nmanmade}{RGB}{213, 213, 213}
\definecolor{nvegetation}{RGB}{0, 175, 0}

\begin{table*}[t] %
	\centering
	\footnotesize
	\setlength{\tabcolsep}{0.003\linewidth}
	\renewcommand\arraystretch{1.0}
	\caption{
		% \textbf{
		Results on nuScenes-SurroundOcc~\cite{surroundocc2023iccv} Validation Set
		% }
	}
	\label{tab:surroundocc}
%	\resizebox{\textwidth}{!}{
\begin{threeparttable}
		\begin{tabular}{lr|cc|cc|cccccccccccccccc|c}
			\toprule
			Method
			& Venue & Represent. & Num.
			& IoU
			& mIoU
			& \rotatebox{90}{\textcolor{nbarrier}{$\blacksquare$} barrier}
			& \rotatebox{90}{\textcolor{nbicycle}{$\blacksquare$} bicycle}
			& \rotatebox{90}{\textcolor{nbus}{$\blacksquare$} bus}
			& \rotatebox{90}{\textcolor{ncar}{$\blacksquare$} car}
			& \rotatebox{90}{\textcolor{nconstruct}{$\blacksquare$} const. veh.}
			& \rotatebox{90}{\textcolor{nmotor}{$\blacksquare$} motorcycle}
			& \rotatebox{90}{\textcolor{npedestrian}{$\blacksquare$} pedestrian}
			& \rotatebox{90}{\textcolor{ntraffic}{$\blacksquare$} traffic cone}
			& \rotatebox{90}{\textcolor{ntrailer}{$\blacksquare$} trailer}
			& \rotatebox{90}{\textcolor{ntruck}{$\blacksquare$} truck}
			& \rotatebox{90}{\textcolor{ndriveable}{$\blacksquare$} drive. surf.}
			& \rotatebox{90}{\textcolor{nother}{$\blacksquare$} other flat}
			& \rotatebox{90}{\textcolor{nsidewalk}{$\blacksquare$} sidewalk}
			& \rotatebox{90}{\textcolor{nterrain}{$\blacksquare$} terrain}
			& \rotatebox{90}{\textcolor{nmanmade}{$\blacksquare$} manmade}
			& \rotatebox{90}{\textcolor{nvegetation}{$\blacksquare$} vegetation}
%			& \rotatebox{90}{barrier}
%			& \rotatebox{90}{bicycle}
%			& \rotatebox{90}{bus}
%			& \rotatebox{90}{car}
%			& \rotatebox{90}{const. veh.}
%			& \rotatebox{90}{motorcycle}
%			& \rotatebox{90}{pedestrian}
%			& \rotatebox{90}{traffic cone}
%			& \rotatebox{90}{trailer}
%			& \rotatebox{90}{truck}
%			& \rotatebox{90}{drive. suf.}
%			& \rotatebox{90}{other flat}
%			& \rotatebox{90}{sidewalk}
%			& \rotatebox{90}{terrain}
%			& \rotatebox{90}{manmade}
%			& \rotatebox{90}{vegetation}
			& FPS
			\\
			\midrule

			TPVFormer~\cite{tpvformer2023cvpr} & CVPR23 & Planes & 3 & {30.9} & 17.1 & 16.0 & \ 5.3 & 23.9	& 27.3 & \ 9.8 & \ 8.7 & \ 7.1 & \ 5.2 & 11.0 & 19.2 & {38.9} & {21.3} & {24.3} & {23.2} & 11.7 & 20.8 & 2.9 \\

			SurroundOcc~\cite{surroundocc2023iccv} & ICCV23 & Dense Vox & 640K & {31.5} & {20.3}  & {20.6} & {11.7} & 28.1 & 30.9 & {10.7} & {15.1} & \second{14.1} & \third{12.1} & {14.4} & 22.3 & 37.3 & {23.7} & {24.5} & {22.8} & {14.9} & {21.9} & 3.3 \\
			
			Co-Occ~\cite{coocc2024ral}  & RAL24 & Sparse Vox & --- & 30.0 & 20.3 & \third{22.5} & 11.2 & \third{28.6} & 29.5 & \ 9.9 & 15.8 & 13.5 & 8.7 & 13.6 & 22.2 & 24.9 & 23.1 & 24.2 & 24.1 & \third{18.0} & 24.8 & ---\\
			
			HybridOcc~\cite{hybridocc2024ral} & RAL24 & Sparse Vox & --- & 33.1 & 21.4 & 22.3 & 12.1 & \second{29.8} & \second{32.3} & 10.9 & 16.3 & \second{14.1} & \second{12.7} & {14.6} & \third{24.0} & 40.4 & 23.7 & 26.2 & 24.5 & 15.2 & 22.6 & 1.9\\

			GaussianFormer-2~\cite{gaussianformer22025cvpr} & CVPR25 & Gaussians & 12.8K & 30.6 & {20.0} & 20.2 & {13.0} & {27.6} & {30.2} & {11.2} & {15.3} & {12.6} & \ {9.6} & {13.3} & {22.3} & {39.7} & {23.5} & {25.6} & {23.2} & 12.3 & 20.7 & 2.8 \\

			QuadricFormer~\cite{quadformer2025nips}  & NeurIPS25 & SQuadrics & 1600      & {31.2} & {20.1} & 19.6 & \third{13.1} & 27.3 & 29.6 & {11.3} & {16.3} & 12.7 & \ 9.2 & 12.5 & 21.2 & {40.2} & {24.3} & {25.7} & {24.2} & 13.0 & 21.9 & {6.2}\\
			
			GaussianWorld~\cite{zuo2025gaussianworld}   & CVPR25 & Gaussians & 25.6K & {33.4} & {22.1} & {21.4} & \best{14.1} & 27.7 & {31.8} & {13.7} & \second{17.4} & \third{13.7} & {11.5} & \second{15.1} & {23.9} & {43.0} & {24.9} & {28.8} & {26.7} & {15.7} & {24.7} & 4.4 \\
			
			VG3S~\cite{vg3s2026iros} & IROS26 & Gaussians & 25.6K & 34.4 & 21.5 & 20.8 & 12.4 & 28.1 & 29.5 & 11.9 & 15.6& 12.4 & 10.6 & \third{14.7} & 21.7 & 42.4 & 26.4 & 28.1 & 26.6 & 17.5 & 25.8 & ---\\
			
			S2GO-Base*~\cite{s2go2026iclr} & ICLR26 & Gaussians & 3600 & \third{35.5} & \third{22.7} & 21.9 & \second{13.4} & 27.5 & \third{32.1} & \third{14.9} & 15.3 & 12.9 & {11.8} & 13.4 & \third{24.0} & \second{46.9} & \third{29.1} & \third{30.3} & \third{29.1} & 14.7 & \third{26.4} & \best{19.6} \\
			
			SuperOcc-L*~\cite{superocc2026icra} & arXiv26 & SQuadrics & 14.4K & \second{38.1} & \best{24.6} & \second{23.9} & {12.4} & \second{29.8} & \best{33.6} & \best{17.4} & \best{17.7} & \best{15.0} & \best{14.8} & 13.4 & \best{26.3} & \best{48.2} & \second{29.2} & \best{32.2} & \second{30.2} & \second{18.7} & \second{30.0} & \second{12.7} \\

			\midrule
			GNFormer (ours) & Ours & GNF & 1024 & \best{38.9} & \second{24.3} & \best{24.4} & 11.2 & \best{31.4} & {31.9} & \second{15.2} & \third{17.0} & 11.8 & 9.8 & \best{17.1} & \second{25.3} & \third{46.0} & \best{30.3} & \second{31.7} & \best{30.4} & \best{23.9} & \best{30.9} & \third{8.7}\\
			\bottomrule
	\end{tabular}
%	}
{
* denotes using sequential inputs and streaming model structures; all others use single-frame input.
FPS results are measured on an RTX 4090 GPU.
}
\end{threeparttable}
\end{table*}

\subsection{Implementation Details}\label{exp:imple}

For all datasets, we set the number of decoder blocks per refinement $N_{\mathrm{blk}}$ to $3$, the number of refinements $B$ to $4$,
and the Gaussian regularization weight $\lambda$ to $0.2$.
We use the AdamW optimizer~\cite{adamw2019iclr} with a learning rate of $2\times10^{-4}$, $\beta_1=0.9$, $\beta_2=0.999$,
and a weight decay of $10^{-2}$. The learning rate of the encoder is set to $2\times10^{-5}$.
The initial quaternions of Gaussians are set to $(1,0,0,0)$.
The support radius $\tau$ is set to $3$ for feature aggregation~\Cref{eq:agg_local} and $6$ for Gaussian regularization~\Cref{eq:flm_loss}.
We initialize the ViT-B~\cite{dinov22023} encoder with the weights from DepthAnythingV2~\cite{dav22024nips} and the ViT-L encoder with the weights from VGGT~\cite{vggt2025wang}.
Dataset-specific hyperparameters are summarized in \Cref{tab:datasets}.

The refinement head $\operatorname{MLP_r}$ consists of two blocks, each containing a linear layer, ReLU, and layer normalization, followed by a linear layer.
The Gaussian feature head $\operatorname{MLP_f}$ consists of two linear layers with ReLU activations, followed by layer normalization and a linear layer.
The semantic decoder that produces $\mbf{c}_i$ follows the occupancy decoder in~\cite{conet2020eccv} and comprises five residual layers.

\subsection{Main Results}\label{exp:results}

\subsubsection{On Occ-ScanNet}
The results are reported in \Cref{tab:exp_occscannet}.
Our GNFormer achieves the best results over all metrics.
By utilizing GNF, GNFormer achieves the fastest inference speed, running at 30.8 FPS, and uses 256 Gaussians, only $1/4$ of the primitives used by FLM-Occ~\cite{gcchen2026flmocc}.

\subsubsection{On SSCBench-KITTI-360}
The results are reported in \Cref{tab:exp_kitti}.
Since 85\% of the voxels in SSCBench-KITTI-360~\cite{li2024sscbench} are labeled as \emph{unknown}, FPS are heavily affected by how these voxels are handled and is therefore not a reliable measure of inference efficiency.
Following prior works, we do not report FPS on this dataset.
GNFormer achieves the best overall IoU and mIoU.
Although it obtains the highest class-wise IoU for most categories, there are still six out of 18 classes where GNFormer does not rank first.
These classes are challenging for all methods, with IoUs generally below 10\%.
Among them, the \emph{pole} class is particularly difficult for Gaussian-based methods;
nevertheless, GNFormer still outperforms other Gaussian-based approaches on this class.

\begin{table*}[t]
    \centering
    \footnotesize
    \setlength{\tabcolsep}{0.008\linewidth}
    \renewcommand\arraystretch{1.0}
    \caption{
%        \textbf{
            Ablation Study and Latency Analysis of GNFormer.
%            }
    }
    \label{tab:abla_latency}
\begin{threeparttable}
% \resizebox{\linewidth}{!}{%
\begin{tabular}{r|ccc|ccccc|rrr|rrcr}
\toprule
\multirow{2}{*}{\rotatebox{90}{Index}}
& \multicolumn{3}{c|}{Representation}
& \multicolumn{5}{c|}{Architecture}
& \multicolumn{3}{c|}{Metrics $\uparrow$}
& \multicolumn{4}{c}{Latency (ms) $\downarrow$} \\
\cline{2-16}
& $\operatorname{MLP_s}$ & Gau. Reg. & \#Gau
& \#Tokens/Img & $N_\text{blk}$ & $B$ & BiCA & ViT
& IoU & mIoU & FPS
& Encoder & Decoder & GNF-to-Voxel & Full \\
\midrule
B
& \checkmark & \checkmark & 1024
& 1590 & 3 & 4 & \checkmark & B
& 37.4 & 23.0 & 8.3
& 44 & 55 & 21 & 120 \\
\midrule
1
& \cellcolor{yellow}{} & \checkmark & 1024
& 1590 & 3 & 4 & \checkmark & B
& \sub{-6.3} & \sub{-3.5} & \sub{-1.1}
& 43 & 54 & \sub{42} & \sub{138} \\

2
& \checkmark & \cellcolor{yellow}{} & 1024
& 1590 & 3 & 4 & \checkmark & B
& \sub{-3.5} & \sub{-1.9} & \sub{-2.5}
& 42 & 54 & \sub{74} & \sub{171} \\

3
& \checkmark & \checkmark & \cellcolor{yellow}{\ 512}
& 1590 & 3 & 4 & \checkmark & B
& \sub{-0.8} & \sub{-0.8} & \add{+1.2}
& 43 & \add{44} & \add{18} & \add{106} \\

4
& \checkmark & \checkmark & \cellcolor{yellow}{2048}
& 1590 & 3 & 4 & \checkmark & B
& \add{+0.3} & \add{+0.1} & \sub{-2.5}
& 43 & \sub{81} & \sub{24} & \sub{148} \\
\midrule
5
& \checkmark & \checkmark & 1024
& \cellcolor{yellow}{\ 777} & 3 & 4 & \checkmark & B
& \sub{-0.5} & \sub{-1.1} & \add{+5.0}
& \add{22} & \add{33} & 20 & \add{75} \\

6
& \checkmark & \checkmark & 1024
& \cellcolor{yellow}{2405} & 3 & 4 & \checkmark & B
& \add{+0.3} & \add{+0.3} & \sub{-2.4}
& \sub{70} & \sub{79} & 20 & \sub{169} \\

7
& \checkmark & \checkmark & 1024
& 1590 & \cellcolor{yellow}{2} & 4 & \checkmark & B
& \sub{-0.1} & \sub{-0.1} & \add{+1.6}
& 44 & \add{36} & 21 & \add{101} \\

8
& \checkmark & \checkmark & 1024
& 1590 & \cellcolor{yellow}{4} & 4 & \checkmark & B
& \add{+0.6} & \sub{-0.2} & \sub{-1.0}
& 44 & \sub{74} & 20 & \sub{137} \\

9
& \checkmark & \checkmark & 1024
& 1590 & 3 & \cellcolor{yellow}{3} & \checkmark & B
& \sub{-0.4} & \sub{-0.5} & \add{+1.4}
& 43 & \sub{40} & 20 & \sub{103} \\

10
& \checkmark & \checkmark & 1024
& 1590 & 3 & \cellcolor{yellow}{5} & \checkmark & B
& \add{+0.4} & \add{+0.1} & \sub{-0.5}
& 43 & \sub{67} & 19 & \sub{128} \\

11
& \checkmark & \checkmark & 1024
& 1590 & 3 & 4 & \cellcolor{yellow}{} & B
& \sub{-0.2} & \sub{-0.9} & \add{+3.6}
& 42 & \add{22} & 20 & \add{84} \\
\midrule
L
& \checkmark & \checkmark & 1024
& 1590 & 3 & 4 & \checkmark & \cellcolor{yellow}{L}
& \add{+1.5} & \add{+1.3} & \sub{-4.0}
& \sub{130} & \sub{84} & 19 & \sub{232} \\
\bottomrule
\end{tabular}%
{
``BiCA'' denotes bidirectional cross-attention.
The metrics are reported as changes relative to Row B.
Latency and FPS are measured on a single RTX 3090.
Yellow highlights configuration changes.
Red and green indicate performance degradation and improvement, respectively.
}
% }
\end{threeparttable}
\end{table*}

\subsubsection{On nuScenes}
The results are reported in \Cref{tab:surroundocc}.
GNFormer achieves the best IoU and the second-best mIoU.
Its class-wise IoUs are comparable to those of S2GO~\cite{s2go2026iclr} and SuperOcc~\cite{superocc2026icra}, both of which are sequential methods that take multiple frames as input.
GNFormer still struggles on several minor classes, such as \emph{bicycle}, \emph{pedestrian}, and \emph{traffic cone}.
In terms of efficiency, GNFormer runs at 8.7 FPS, compared with 19.6 FPS for S2GO and 12.7 FPS for SuperOcc.
Nevertheless, compared with single-frame methods, GNFormer achieves the best performance on most classes and also attains the fastest inference speed.

\subsection{Ablation Studies on Representation and Architecture}

\Cref{tab:abla_latency} reports the results of ablation studies on the representation and architecture on the nuScenes-SurroundOcc benchmark~\cite{surroundocc2023iccv}.
Row B is the smaller baseline, while Row L is the final GNFormer configuration in \Cref{tab:surroundocc}.

The results show that the representation and the encoder have much more impacts on the performance.
Rows 1 and 2 ablate the two core designs of GNF, neural decoding through $\operatorname{MLP_s}$ and Gaussian regularization, respectively.
Removing the neural decoding decreases IoU and mIoU by 6.3 and 3.5 percentage points, demonstrating the accuracy benefit by using expressive neural decoding.
Without Gaussian regularization, IoU and mIoU decrease by 3.5 and 1.9 points, showing the importance of Gaussian spatial allocation.
Rows 3 and 4 further show that reducing the number of Gaussians from 1,024 to 512 incurs only 0.8-point drops in both metrics, whereas doubling it to 2,048 yields gains of only 0.3 and 0.1 points.
GNF achieves better results without relying on a large primitive budget.

Among the architectural choices, ViT-L provides the largest gain, improving IoU and mIoU by 1.5 and 1.3 points, respectively.
Removing bidirectional cross-attention reduces mIoU by 0.9 points, whereas varying the image tokens, decoder blocks, or refinement stages produces only modest changes.
As reported in \Cref{tab:abla_enc}, the evaluated encoder initializations differ by at most 0.4 and 0.3 percentage points in IoU and mIoU, respectively.

\subsection{Efficiency Analysis}

\begin{table}[t] %
    \centering
    \footnotesize
    \renewcommand\arraystretch{1.0}
    \caption{
        % \textbf{
            Ablation Study of Encoder Initialization
        % }
    }
    \label{tab:abla_enc}
\begin{tabular}{l|c|c}
\toprule
ViT Initialization & IoU & mIoU \\
\midrule
DepthAnythingV2~\cite{dav22024nips}  & 38.5 & 24.0\\
$\pi^3$-X~\cite{pi32026iclr} & 38.6 & 24.0\\
DINOv2~\cite{dinov22023} & 38.7 & 24.3 \\
VGGT~\cite{vggt2025wang} & 38.9 & 24.3\\
\bottomrule 
\end{tabular}
\end{table}

\begin{table}[t]
\centering
\footnotesize
\setlength{\tabcolsep}{0.006\linewidth}
\renewcommand\arraystretch{1.0}
	\caption{
		% \tbf{
				Aggregation Efficiency
		% }
	}
\label{tab:aggregation-efficiency}
\begin{threeparttable}
\begin{tabular}{l|cc|cc}
\toprule
& \multicolumn{2}{c|}{Density Agg. / Occupied} &
	  \multicolumn{2}{c}{Feat. Agg. / Full Grid} \\
% \cmidrule(lr){2-3} \cmidrule(lr){4-5}
\cline{2-3} \cline{4-5}
Method & Mean (ms) & Peak (MiB) & Mean (ms) & Peak (MiB) \\
\midrule
Naive PyTorch  & 60.13 & \textbf{37.25} & 998.60 & \textbf{346.26} \\
GaussianFormer-2~\cite{gaussianformer22025cvpr} & \ 1.14 & 91.86 & \ \ 2.55 & 576.16 \\
SuperOcc~\cite{superocc2026icra}       & \ 0.85 & 55.60 & \textbf{\ \ 2.12} & 526.28 \\
Ours (\cref{sec:pgf})     & \textbf{\ 0.43} & \textbf{37.25} & \ \ 6.47 & 353.27 \\
\bottomrule
\end{tabular}
Densities and features are aggregated over occupied voxels and the full grid, respectively.
The complexity of \emph{Naiev PyTorch} is $\mathcal{O}(PM)$ for \cref{eq:agg}.
Other methods accelerate \cref{eq:agg_local} while ours leverages \cref{alg:pgf}.
\end{threeparttable}
\end{table}

Beyond their accuracy gains, the two core components of GNF also reduce inference latency (\Cref{tab:abla_latency}).
Introducing $\operatorname{MLP_s}$ halves GNF-to-voxel conversion time from 42 to 21 ms.
Its greater representational capacity allows occupancy to be modeled with smaller Gaussian supports, reducing the decoding range around each primitive.
Gaussian regularization yields an even larger reduction, from 74 to 21 ms, by concentrating Gaussian supports in occupied regions and avoiding unnecessary decoding in empty space.
Together, these complementary mechanisms reduce the overall range of neural decoding while improving occupancy accuracy.

Latency changes from the architecture variants largely follow their computational scale, with the image token count and bidirectional cross-attention having the greatest impact.

\Cref{tab:aggregation-efficiency} shows that projection-based pairing efficiently performs the local computations required by GNF, substantially accelerating both Gaussian regularization and feature aggregation over naive point--Gaussian enumeration.

\section{Conclusion}
We presented GNF, a novel 3D representation for compactly parameterizing a semantic occupancy grid.
It leverages Gaussian regularization and the local support property of Gaussians to restrict neural decoding around occupied regions.
Together with a neural decoder, GNF achieves higher accuracy than pure Gaussians while reducing primitive budget, computation, and memory costs.
Leveraging GNF, we were able to develop GNFormer, a novel camera-based occupancy prediction method that leverages Transformer blocks predicts GNFs for accurate and efficient inference.
Thanks to its streamlined architecture, it can be flexibly trained for different scenarios and camera setups.
Across Occ-ScanNet, SSCBench-KITTI-360, and nuScenes-SurroundOcc, GNFormer achieves leading IoU, mIoU, and inference efficiency among single-frame methods with only 256 Gaussians indoors and 1,024 outdoors, demonstrating its effectiveness and efficiency for robot perception.

Despite its strong performance among single-frame methods, GNFormer relies solely on single-frame inputs and therefore requires a larger model to achieve high accuracy. As a result, its inference speed remains lower than that of recent streaming-based methods.
In addition, due to annotation discrepancies across datasets, we train separate models for indoor and outdoor scenarios following prior works.

\bibliographystyle{IEEEtran}
\bibliography{ref}

\end{document}